\documentclass[letterpaper]{article} 
\usepackage{aaai19}  
\usepackage{times}  
\usepackage{helvet}  
\usepackage{courier}  
\usepackage{url}  
\usepackage{graphicx}  
\usepackage{subcaption}
\usepackage{amsmath}
\usepackage{amsfonts}
\usepackage[ruled,linesnumbered]{algorithm2e}
\usepackage{algpseudocode}
\usepackage{multirow}
\usepackage{siunitx}
\usepackage{bbm}
\usepackage{caption}
\usepackage{bm}
\usepackage{fixmath}
\begin{document}

\title{SAFE: A Neural Survival Analysis Model for Fraud Early Detection}
\author{Panpan Zheng,\thanks{Equal contribution.}  Shuhan Yuan,\footnotemark[1]  Xintao Wu \\ University of Arkansas, Fayetteville, AR, USA \\ \{pzheng, sy005, xintaowu\}@uark.edu}

\maketitle

\begin{abstract}
Many online platforms have deployed anti-fraud systems to detect and prevent fraudulent activities. However, there is usually a gap between the time that a user commits a fraudulent action and the time that the user is suspended by the platform. How to detect fraudsters in time is a challenging problem. Most of the existing approaches adopt classifiers to predict fraudsters given their activity sequences along time. The main drawback of classification models is that the prediction results between consecutive timestamps are often inconsistent. In this paper, we propose a survival analysis based fraud early detection model, SAFE, which maps dynamic user activities to survival probabilities that are guaranteed to be monotonically decreasing along time. SAFE adopts recurrent neural network (RNN) to handle user activity sequences and directly outputs hazard values at each timestamp, and then, survival probability derived from hazard values is deployed to achieve consistent predictions. Because we only observe the user suspended time instead of the fraudulent activity time in the training data, we revise the loss function of the regular survival model to achieve  fraud early detection. Experimental results on two real world datasets demonstrate that SAFE outperforms both the survival analysis model and  recurrent neural network model alone as well as state-of-the-art fraud early detection approaches.
\end{abstract}

\section{Introduction}
Due to the openness and anonymity of the Internet, online platforms (e.g., online social media or knowledge bases) attract a large number of malicious users, such as vandals, trolls, and sockpuppets. These malicious users impose severe security threats to online platforms and their legitimate participants. For example, the fraudsters on Twitter can easily spread fake information or post harmful links on the platform. To protect legitimate users, most web platforms deploy tools to detect fraudulent activities and further take actions (e.g., warning or suspending) against those malicious users. However, there is usually a gap between the time that fraudulent activities occur  and the time that response actions are taken. Training datasets collected and used for building new detection algorithms often contain the labeled information about when users are suspended instead of when users take fraudulent actions. For example, using twitter streaming API and crawler can easily collect the suspended time information of fraudsters in addition to a variety of dynamically changing features (e.g., the number of posts or the number of followers). However, there is no ground truth about when fraudulent activities occur from the collected data. Hence, the algorithms trained on such datasets cannot achieve in-time or even early detection if they do not take into consideration the gap between suspended time and fraudulent activity time. In this work, we aim to develop effective fraud early detection algorithms  over such training data that contains time-varying features and late response labels.

\begin{figure}[htb]
    \centering
  	\includegraphics[width=0.4\textwidth, keepaspectratio]{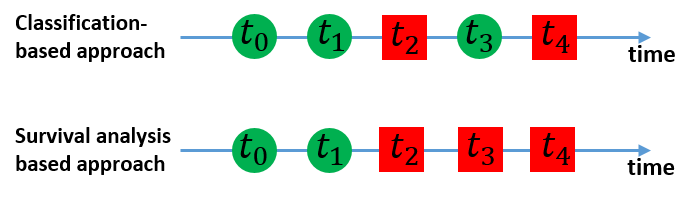}
  	\caption{Comparison of the survival analysis-based approach and classification-based approach for fraud early detection. Red square indicates that the user is predicted as fraudsters at time $t$ while the green circle indicates the user is predicted as normal.}
  	\label{fig:sv_class}
\end{figure}

Fraud early detection has attracted increasing attention in the research community \cite{Kumar2015Vews,Yuan2017Wikipedia,Wu2017Gleaning,Zhao2015Enquiring}.
The existing approaches for fraud early detection are usually based on classification models (e.g., neural network, SVM). Given a sequence of user activities that contain intermittent fraudulent activities, the prediction at each timestamp from the built classifier is often independent to each other. Hence, these classification models tend to make inconsistent and ad-hoc predictions along the time.  Figure \ref{fig:sv_class} shows an illustrative example. A user takes a fraudulent action at time $t_2$,  the classification model predicts the user as a fraudster at $t_2$ and $t_4$ but as normal user at $t_3$. This is because the prediction probabilities between consecutive timestamps do not have any relations.

In this work, we propose to use the survival analysis \cite{Klein2006Survival} to achieve consistent predictions along the time. Survival analysis models the time until an event of interest occurs and incorporates two types of information: 1) whether an event occurs or not, and 2) when the event occurs. In survival analysis,  \textit{hazard rate} and \textit{survival probability} are adopted to model event data. The hazard rate at time $t$ indicates the instantaneous rate at which events occur, given no previous event whereas the survival probability indicates the probability that a subject will survive past time $t$.

In the fraud detection scenario, the event is that a fraudster is suspended by the platform. We use the survival function, which is monotonically decreasing, to model the likelihood of being fraudster for a given user based on his observed activities. Hence, unlike the classification model that makes ad-hoc predictions, the survival model can keep track of user survival probabilities over time and provide consistent prediction. When deployed, the survival analysis model can easily calculate the survival probability of a new user at each timestamp based on his activities and predict the user as a fraudster when the survival probability is below some threshold.

However, it is nontrivial to adopt survival analysis for fraud detection. Traditional survival analysis
models often assume a specific parametric distribution of underlying data. However, it is generally unknown which distribution fits well in fraud detection scenarios. We need a model to handle the features of user activity sequences (time-varying covariates) and further capture general relationships between the survival time distribution and time-varying covariates. To tackle this challenge, we develop a neural Survival Analysis model for Fraud Early detection (SAFE) by combining the recurrent neural network (RNN) with the survival analysis model. SAFE adopts RNN to handle time-varying covariates as inputs and predicts the evolving hazard rate given the up-to-date covariates at each timestamp. RNN can capture the non-linear relations between the hazard rates and time-varying covariates and does not assume any specific survival time distributions. Moreover, to tackle the challenge due to the gap between suspended time (reported in training data) and fraudulent activity time (unavailable in training data), we revise the loss function of the regular survival model. In particular, SAFE is trained to intentionally increase the hazard rates of fraudsters before they are suspended  and decrease  the hazard rates of normal users.

The contributions of this work are as follows. First, it is the first work to adopt survival analysis for fraud detection. Different from classification models, our approach achieves consistent predictions along the time.  Second, our revised survival model is designed for the training data with late response labels and can achieve fraud early detection. Third, instead of assuming any particular survival time distributions, we propose the use of RNN to learn the hazard rates of users from user activities along time and do not assume any specific distribution.  Fourth, we conduct evaluations over two real-world datasets and our model outperform state-of-the-art fraud detection approaches.

\section{Related Work}

{\bf \noindent Survival analysis:}
Survival analysis is to analyze and model the data where the outcome is the time until the occurrence of an event of interest \cite{Wang2017Machine}.  In survival analysis, the occurrence of an event is not always observed in an observation window, which is called \textit{censored}.

Survival analysis is a widely-used tool in health data analysis \cite{Liu2018Early,Ranganath2016Deep,Yu2011Learning} and has been applied to various application fields, such as students dropout time \cite{Ameri2016Survival}, web user return time \cite{Jing2017Neural,Du2016Recurrent,Barbieri2016Improving}, and user check-in time prediction \cite{Yang2018SpatioTemporal}. To our knowledge, the survival analysis has not been investigated in the context of fraud detection.

Many approaches have been proposed to make use of censored data as well as the event data. The Cox proportional hazards model (CPH) \cite{Cox1972Regression} is the most widely-used model for survival analysis. CPH is semi-parametric and does not make any assumption about the distribution of event occurrence time. It is typically learned by optimizing a partial likelihood function. However, CPH makes strong assumptions that the log-risk of an event is a linear combination of covariates, and the base hazard rate is constant over time. Some researchers proposed parametric censored models, which assume the event occurrence time follows a specific distribution such as exponential, log-logistic or Weibull \cite{Alaa2017Deep,Ranganath2016Deep,Martinsson2016WtteRnn}. However, it is common that the specific parametric assumptions are not satisfied in real data.

In recent years, researchers adopt neural networks to model the survival distribution \cite{Luck2017Deep,Katzman2018Deepsurv,Lee2018Deephit,Chapfuwa2018Adversarial,biganzoli1998feed}. For example, \cite{Luck2017Deep,Katzman2018Deepsurv} combine the feed-forward neural network with the classical Cox proportional hazard model. Although using the deep neural network can improve the capacity of models, these studies still assume that the base hazard rate is constant. \cite{Lee2018Deephit} transfers the problem of learning the distribution of survival time to a discretized-time classification problem and adopts the deep feed forward neural network to predict the survival time. \cite{Chapfuwa2018Adversarial} adopts a conditional generative adversarial network  to predict the event time conditioned on covariates, which implicitly specifies a time-to-event distribution via sampling. However, the existing models cannot handle the time-varying covariates. In this work, we adopt the RNN to take the time-varying covariates as inputs and fit the time-to-event distribution without making any of the above assumptions.

We also notice that some studies adopt RNN to model the time-to-event distributions. Those studies mainly focus on modeling the recurrent event instead of the terminated event. For example, \cite{Du2016Recurrent,Jing2017Neural,Grob2018Recurrent} adopt RNN to model the web user return times, which focus on the recurrent event data other than the censored data. Hence, RNN is to capture the gap time between user active sessions. Moreover, unlike the existing work that focuses on ``just-in-time'' prediction, we adapt the survival analysis for fraud early detection in the scenario where training data contains late response labels.

{\bf \noindent Fraud early detection:}
The misleading or fake information spread by malicious users could lead to catastrophic consequences because the openness of online social media enables the information to be spread in a timely manner. Therefore, detecting fake information or malicious users is a critical research topic \cite{Ying2011Spectrum,Yuan2017SpectrumBased,DBLP:journals/jiis/WuWLZ13,Manzoor2016Fast,Kumar2018False}. In recent years, extensive studies focus on the rumor early detection \cite{Wu2017Gleaning,Zhao2015Enquiring}. Besides early detecting the fake information, early detecting the malicious users who create the fake information is also important. \cite{Kumar2015Vews,Yuan2017Wikipedia} aim to early detect vandals in Wikipedia. All the existing approaches adopt classification models for fraud early detection. In this work, we combine the survival analysis with RNN to predict whether a user is a fraudster.

\section{Preliminary: Survival Analysis}
Survival analysis models the time until an event of interest occurs. Compared with the common regression models, in a survival analysis experiment, we may not always be able to observe event occurrence from start to end due to missing observation or a limited observation window size. For example, in health data analysis, the time of death can be missing  in some patient records. Such phenomenon is called \textit{censoring}. In this work, we focus on two types of censoring: 1) an \textit{uncensored} sample indicates the event is observed; 2) a \textit{right censored} sample indicates the event is not observed in the observation window but we know it will occur later.

Survival time $T$ is a continuous random variable representing the waiting time until the occurrence of an event, with the probability density function $f(t)=\lim_{dt \to 0} \frac{P\{t \leq T < t + dt \}}{dt}$ and the cumulative distribution function $F(t)= P(T < t)=\int_{0}^{t} f(x) dx$.

The \textit{survival function} $S(t)$ indicates the probability of the event having not occurred by time $t$:
\begin{equation}
    S(t)=P(T \geq t)= 1-F(t)=\int_{t}^{\infty} f(x) dx.
\end{equation}

The \textit{hazard function} $\lambda(t)$ refers to the instantaneous rate of occurrence of the event at time $t$ given that the event does not occur before time $t$:
\begin{equation}
\label{eq:original_hazard}
  \lambda \mathit{(t)} = \lim_{dt \to 0}\frac{P\{\mathit{t \leq T < t + dt | T \geq t} \}}{dt}
  = \frac{\mathit{f(t)}}{\mathit{S(t)}}.
\end{equation}

Additionally, $S(t)$ is associated with $\lambda(t)$ by
\begin{equation}
  S(t) = e^{-\int_{0}^{t} \lambda (x)dx}.
\end{equation}

{\bf\noindent Discrete time.}
In many cases, the observation time is discrete (seconds, minutes or days). When $T$ is a discrete variable,  we denote $t$  a timestamp index and have the discrete expression:
\begin{align}
	S_t & = P\{T \geq t\} = \sum_{k=t}^\infty f_k, \label{eq:survival_func_disc}\\
	\lambda_t & = P\{T=t|T \geq t\} = \frac{f_t}{S_t}, \label{eq:hazard}  \\
	S_t & = e^{-{\sum_{k=1}^{t} \lambda_k}} \label{eq:lambda_survive}.
\end{align}

{\bf\noindent Likelihood function.}
Given a training dataset with $N$ samples where each sample $i$ has an aggregated covariate $\mathbf{x}^i$, a last-observed time $t^i$, and an event indicator $c^i$, the survival model adopts maximum likelihood to estimate the hazard rate and the corresponding survival probability. If a sample $i$ has the event ($c^i=1$), the likelihood function seeks to make the predicted time-to-event equal to the true event time $t^i$, i.e., maximizing $P\{T = t^i\}$; if a sample $i$ is censored ($c^i=0$), the likelihood function aims to make the sample survive over the last-observed time $t^i$, i.e., maximizing $P\{T \geq t^i\}$. The joint likelihood function for a sample $i$ is:
\begin{equation}
\label{eq:regular_likelihood}
	P\{T = t^i\}^{c^i} \cdot P\{T \geq t^i\}^{1-c^i} = f(t^i)^{c^i}S(t^i)^{1-c^i}.
\end{equation}
The negative log-likelihood function for a sample $i$ can be written as:
\begin{equation}
\begin{split}
\label{eq:loss_sv_idv}
  \ell^i_r &= -[c^i \ln(P\{T = t^i\}) + (1-c^i) \ln(P\{T \geq t^i\})] \\
       &= \big(\sum_{t=1}^{t^i}\lambda_t \big) - c^i \cdot \ln (e^{\lambda_{t^i}}-1),
\end{split}
\end{equation}
where $\lambda_t = \lambda(t|\mathbf{x}^i_{t};\theta)$ is the conditional hazard rate given covariate $\mathbf{x}$ with parameters $\theta$.

The overall loss function over the whole training data is:
\begin{equation}
\label{eq:loss_sv}
  \mathcal{L}_r = \sum_{i=1}^N \ell^i_r 
  = \sum_{i=1}^N \Big[\big(\sum_{t=1}^{t^i}\lambda_t\big) - c^i \cdot \ln \big(e^{\lambda_{t^i}}-1 \big) \Big].
\end{equation}
The survival analysis models learn the relationship between the covariate $\mathbf{x}^i$ and the survival probability $S(t)$ by optimizing parameters $\theta$ to estimate $\lambda_t$.

\section{SAFE: A Neural Survival Analysis Model for Fraud Early Detection}

In the fraud detection scenario, \textit{event of interest} refers to users being suspended by platforms; then, \textit{survival time} corresponds to the length of time that a user is active. Hence, users who are suspended in the observation window are event samples; users who are not suspended are right-censored samples.

\subsection{Problem Statement}

Let $\mathcal{D} = \{(\mathbf{x}^{i}, c^{i}, t^i)\}_{i=1}^N$ denote a set of training triplets, where $\mathbf{x}^{i} = (\mathbf{x}_1^{i}, \mathbf{x}_2^{i}, \cdots, \mathbf{x}_{t^i}^i)$ indicates the sequence data of user $i$; $c^{i}$ indicates whether the user $i$ is suspended ($c^i=1$) or un-suspended ($c^i=0$) in the observation window; $t^i$ denotes the time when the user $i$ is suspended by the platform or the last-observed time for an un-suspended user; $N$ denotes the size of the dataset. We consider the problem of detecting fraudsters in a timely manner. Because $t^i$ is the suspended time by the platform instead of the time of committing malicious activities, we require the detected time earlier than the suspended time $t^i$. The goal of learning is to train a mapping function between time-varying covariates and the survival probabilities, i.e., $S_t=f(\mathbf{x}^{i}_t)$. 
The learned mapping function can be deployed to predict whether a new user is a fraudster at time $t$ based on his activities by comparing the survival probability $S_t$ with a threshold $\tau$. 

\begin{figure}[htb]
    \centering
  	\includegraphics[width=0.35\textwidth, keepaspectratio]{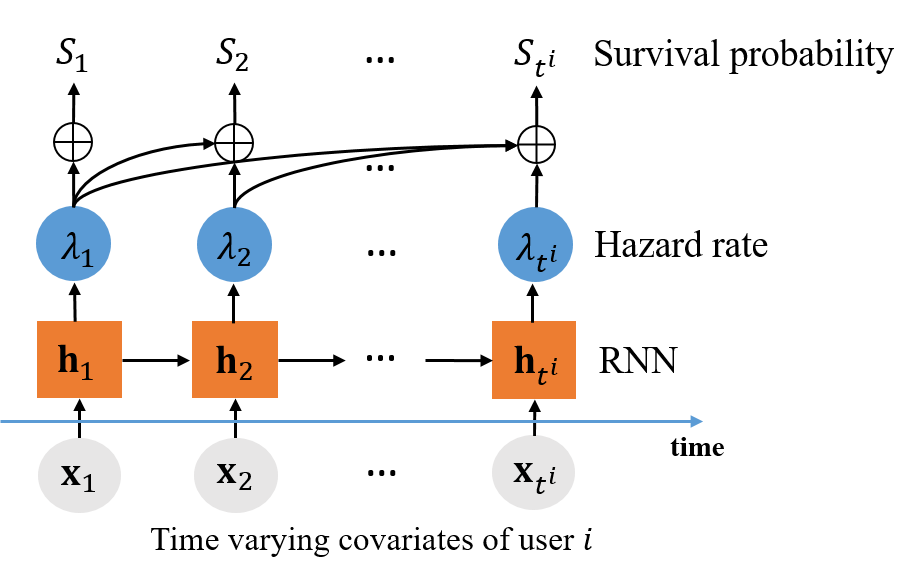}
  	\caption{An RNN-based survival analysis model for fraud early detection}
  	\label{fig:safe}
\end{figure}

\subsection{Model Description}
Figure \ref{fig:safe} describes the basic framework of SAFE. RNN is taken to handle the time-varying covariates and its outputs are hazard rates along time. At timestamp $t$, RNN maintains a hidden state vector $\mathbf{h}_t \in \mathbb{R}^{h}$  to keep track  of users' sequence information from the current input $\mathbf{x}_t$ and all of the previous inputs $\mathbf{x}_{k}$, s.t. $k<t$.

In this work, we adopt the gated recurrent unit (GRU) \cite{Cho2014Learning}, a variant of the traditional RNN, to model the long-term dependency of time-varying covariates. With $\mathbf{x}_t$ and $\mathbf{h}_{t-1}$, the hidden state $\mathbf{h}_t$ is computed by
\begin{equation}
	\mathbf{h}_t = GRU(\mathbf{x}_t, \mathbf{h}_{t-1}).
\end{equation}

As shown in Figure \ref{fig:safe}, at time $t$, hazard rate $\lambda_t$, which indicates the instantaneous rate of a user should be suspended given that the user is still alive at time $t$, is derived from $\mathbf{h}_t$ by
\begin{equation}
	\label{eq:rnn_lambda}
	\lambda_t = softplus(\mathbf{w}_{\lambda} \mathbf{h}_t) = \ln(1+\exp(\mathbf{w}_{\lambda}\mathbf{h}_t)),
\end{equation}
where \textit{softplus($\cdot$)} is deployed to guarantee that hazard rate $\lambda$ is always positive, and $\mathbf{w}_{\lambda}$ is the weight vector of RNN output layer. Note that the softplus function can be replaced by other non-linear functions with positive outputs. 

Based on Equation \ref{eq:lambda_survive}, the survival probability, which indicates the probability of a user having not been suspended until time $t$, can be calculates as $S(t) =e^{-\sum_{k=1}^t \lambda(k)}$. By comparing the survival probability with a threshold $\tau$, we can predict whether a user should be suspended at time $t$. The survival probability $S(t)$ is monotonically decreasing along time, hence we can achieve consistent predictions.

For outputs, unlike previous works \cite{Martinsson2016WtteRnn,Alaa2017Deep}, we do not assume hazard rate $\lambda$ follows one certain parametric distribution, such as Weibull or Poisson, because, in context of fraud early detection, we do not know whether $\lambda$ follows one particular distribution. Instead, SAFE directly outputs $\lambda$ which actually follows a general distribution potentially captured by RNN. We conduct experiments to compare two designs and evaluation results demonstrate SAFE outperforms the design with specific parametric distributions. 

{\bf\noindent Loss function.} The loss function shown in Equation \ref{eq:loss_sv} for traditional survival analysis cannot be used for learning fraud detection model over the training data with late response labels. In our fraud detection scenario, we aim to detect fraudsters as early as possible while let censored users survive over the last-observed time. However, Equation \ref{eq:loss_sv} can let censored users pass over the last-observed time but cannot detect fraudsters as early as possible. 

Aiming to fraud early detection, a simple but non-trivial adaption is performed on Equation \ref{eq:loss_sv} to obtain our early-detection-oriented likelihood function, i.e. Equation \ref{eq:safe_likelihood}. For simplicity, first, we take user $i$ as an example to give the expression of likelihood and loss function, and then show the overall loss function for the whole dataset.

\begin{equation}
\begin{split}
\label{eq:safe_likelihood}
& ~P\{T < t^i\}^{c^i} \cdot P\{T \geq t^i\}^{1-c^i} \\
       &= \big(F(t^i)\big)^{c^i} \cdot S(t^i)^{1-c^i}\\
       &= \big(1-e^{-\sum_{t=1}^{t^i} \lambda_t}\big)^{c^i} \cdot \big(e^{-\sum_{t=1}^{t^i} \lambda_t}\big)^{1-c^i}\\
       &= \big(e^{\sum_{t=1}^{t^i} \lambda_t}-1\big)^{c^i} \cdot e^{-\sum_{t=1}^{t^i} \lambda_t}.
\end{split}
\end{equation}

Compared with the likelihood function of a regular survival model shown in Equation \ref{eq:regular_likelihood}, Equation \ref{eq:safe_likelihood} changes $P\{T = t^i\}^{c^i}$ to $P\{T < t^i\}^{c^i}$. After this adaption, intuitively, we can realize that it does match the fraud early detection: with user $i$ being a fraudster ($c^i=1$), all of hazard rates before $t^i$ will naturally increase as maximizing the term $P\{T < t^i\}$.

Taking the negative logarithm, we could get loss function of user $i$:
\begin{equation}
\begin{split}
\label{eq:loss_early_idv}
  \ell^i=\big(\sum_{t=1}^{t^i}\lambda_t\big) - c^i \cdot \ln (e^{\sum_{t=1}^{t^i}\lambda_{t}}-1).
\end{split}
\end{equation}

Then, given a set of training samples with $N$ users, the overall loss function is defined as:
\begin{equation}
\label{eq:loss_early}
  \mathcal{L} = \sum_{i=1}^N \ell^i 
  = \sum_{i=1}^N \Big[\big(\sum_{t=1}^{t^i}\lambda_t\big) - c^i \cdot \ln \big(e^{\sum_{t=1}^{t^i}\lambda_{t}}-1 \big) \Big].
\end{equation}

Next we illustrate why SAFE is appropriate for fraud early detection. We denote the model trained by the original loss function $\mathcal{L}_r$ (shown in Equation \ref{eq:loss_sv}) as \textbf{SAFE-r}.
For simplicity, instead of two overall loss functions, our following discussions focus on  $\ell^i_r$ and $\ell^i$.  

The first partial derivatives of $\ell^i_r$ and $\ell^i$ w.r.t $\lambda$ are listed as follows:

\begin{equation}
\begin{split}
\label{eq:partial_dev_likelihood}
\frac{\partial \ell_{r}^{i}}{\partial \lambda_t} =
                                    \left\{
                                    \begin{array}{ll}
                                      1  &  0 < t < t^i \\
                                      1 - c^i \cdot \frac{e^{\lambda_{t}}}{e^{\lambda_{t}}-1} &  t = t^i
                                    \end{array}
                                   \right.
\end{split}
\end{equation}

\begin{equation}
\label{eq:partial_dev_likelihood_adp}
\frac{\partial \ell^{i}}{\partial \lambda_t} = 1 - c^i \cdot \frac{e^{\sum_{k=1}^{t^i}\lambda_{k}}}{e^{\sum_{k=1}^{t^i}\lambda_{k}}-1} \quad 0 < t \leq t^i.
\end{equation}

For a fraudster $i$ ($c^i=1$), we can see $\frac{\partial \ell_{r}^{i}}{\partial \lambda_t} = 1 >0, (0<t<t^i)$. It means $\ell_{r}^{i}$ is an increasing function w.r.t $\lambda$ so that $\lambda_t$ ($0 < t < t^i$) is decreasing as minimizing $\ell_{r}^{i}$. Moreover, in accordance with Equation \ref{eq:lambda_survive},  survival probability $S_t$ is increasing with the decrement of $\lambda_t$, which means survival probability $S_t$ is increasing with the minimization of $\ell_{r}^{i}$. That is, instead of detecting the fraudster $i$ before $t^i$, SAFE-r tends to make the fraudster $i$ survive over $t^i$. On the contrary, for SAFE, we can observe that $\frac{\partial \ell^{i}}{\partial \lambda_t} =
1 - c^i \cdot \frac{e^{\sum_{k=1}^{t^i}\lambda_{k}}}{e^{\sum_{k=1}^{t^i}\lambda_{k}}-1}
=1 - \frac{e^{\sum_{k=1}^{t^i}\lambda_{k}}}{e^{\sum_{k=1}^{t^i}\lambda_{k}}-1} < 0$. It means $\ell^{i}$ is a decreasing function w.r.t $\lambda$ so that $\lambda_t$ ($0 < t < t^i$) is increasing as minimizing $\ell^{i}$. Similarly, we can achieve that survival probability $S_t$ is decreasing with $\ell^{i}$ minimized, which implies that SAFE does have a tendency to detect fraudster $i$ before the suspended time $t^i$.

For a censored user $j$ ($c^j=0$), we obtain $\frac{\partial \ell_{r}^{j}}{\partial \lambda_t} = \frac{\partial \ell^{j}}{\partial \lambda_t} = 1$. Both $\ell_{r}^{j}$ and $\ell^{j}$ are increasing functions w.r.t $\lambda$. As minimizing $\ell_{r}^{j}$ or $\ell^{j}$, $\lambda_t$ is becoming smaller. SAFE and SAFE-r both have a tendency to make censored user $j$ survive over the last-observed time $t^j$. 

The above theoretical analysis shows why SAFE can achieve the \textit{fraud early detection} better than SAFE-r. Experimental results in the experiment section also validate this theoretical analysis.

\section{Experiments}

\subsection{Experimental Settings}

\begin{figure}[h]
	\centering 
	\begin{subfigure}{0.23\textwidth}
	\centering
		\includegraphics[width=\linewidth,]{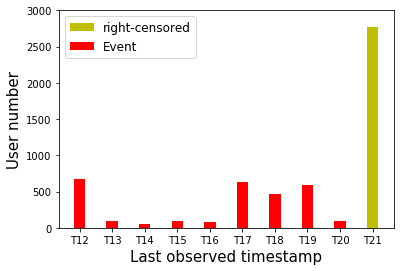}
		\caption{\textit{Twitter}}
		\label{fig:twitter_dist}
	\end{subfigure} \hfil 
	\begin{subfigure}{0.23\textwidth}
	\centering
		\includegraphics[width=\linewidth]{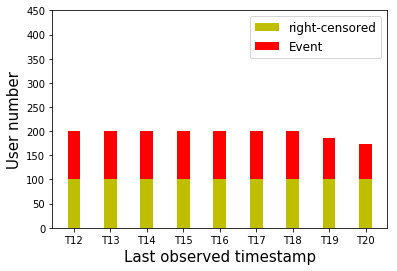}
		\caption{\textit{Wiki}}
		\label{fig:wiki_dist}
	\end{subfigure}
	\caption{The distributions of event and right-censored users over the timestamps on \textit{twitter} and \textit{wiki} datasets}
\end{figure}

{\bf \noindent Datasets.} We conduct our experiments on two real-world datasets:

\begin{itemize}
	\item \textbf{\textit{Twitter}}. We randomly collect 51608 Twitter users on August 13, 2017,  monitor the user statuses every three days until October 13, 2017, and get the data with 21 timestamps. For each user, at each timestamp, the following 5 features are recorded: 1) the number of followers, 2) the number of followees, 3) the number of tweets, 4) the number of liked tweets, and 5) the number of public lists that the user is a member of. During this period, 7790 users ($15.0\%$) are suspended; the remaining 43818 users ($85.0\%$) are still active, i.e., right-censored. We then select suspended users who have the observed timestamps ranging from 12 to 21 and randomly choose the censored users to compose a balanced dataset. To this end, \textit{twitter} consists of 2770 fraudsters and 2770 normal users. We take the change values of five features between two consecutive timestamps as inputs to RNN. Fig.\ref{fig:twitter_dist} details the components of \textit{twitter} involving numbers of event-censored users at different last-observed timestamps.

	\item \textbf{\textit{Wiki}}. We adopt the UMDWikipedia dataset \cite{Kumar2015Vews} to build the \textit{wiki} dataset for early vandal detection. \textit{Wiki} contains 1759 users whose editing sequence lengths are between 12 and 20, where 900 are vandals and 859 are benign users. We collect eight features at each edit for each user: 1) whether the user edits a Wikipedia meta-page, 2) whether the category of the edit page is an empty set, 3) whether the consecutive re-edit is less than one minute, 4) whether the consecutive re-edit is less than three minutes, 5) whether the consecutive re-edit is less than fifteen minutes, 6) whether the current edit page has been edited before, 7) whether the user edits the same page consecutively, and 8)  whether the consecutive re-edit pages have the common category. Fig.\ref{fig:wiki_dist} illustrates the components of \textit{wiki} involving event-censor numbers at different last-observed timestamps. Different from \textit{twitter} where the censored users are in the last timestamp, there are censored users at each timestamp on \textit{wiki}.

\end{itemize}

{\bf \noindent Baselines.} We compare SAFE with the following baselines:

\begin{itemize}
	\item \textbf{SVM} is a classical classifier. Given a user time-varying covariate, we average the sequence of each covariate as input to train the SVM and predict the user types (fraudsters or normal users) at each timestamp at the testing phase.

	\item \textbf{CPH} (Cox proportional hazard model) is a classical survival regression model \cite{Cox1972Regression}. Similar to SVM, we adopt the average covariates of users as input to train CPH and conduct fraud early detection with the first $k$ timestamps. We adopt Lifelines \footnote{\url{https://lifelines.readthedocs.io}} to implement the CPH model.

	\item \textbf{M-LSTM} (Multi-source LSTM) is a classification-based fraud early detection model that adopts LSTM  to capture the information of time-varying covariates and dynamically predict the user type at each timestamp based on the logistic regression classifier \cite{Yuan2017Wikipedia}.

\end{itemize}

{\bf \noindent Hyperparameters.}
SAFE is trained by back-propagation via Adam \cite{Kingma2015Adam} with a batch size of 16 and a learning rate $\num{e-03}$. The dimension of the GRU hidden unit is 32. We randomly divide the dataset into a training set, a validation set, and a testing set with the ratio (7:1:2). The threshold $\tau$ for fraud early detection is set based on the performance on the validation set. We run our approach and all baselines for 10 times and report the mean and standard deviation of each metric. For all the baselines, we use the default parameters provided by the public packages.

{\bf \noindent Evaluation Metrics.}
We use \textit{Precision}, \textit{Recall}, \textit{F1} and \textit{Accuracy} to evaluate the fraud early detection performance of various models given the first $K$-timestamps. For instance, $\textit{Accuracy}@k$ (\textit{k=1,2,3,4,5}) indicates the accuracy given the first K-timestamp inputs. We further report the ``percentage of early detected fraudsters'' to show the portion of correctly early detected fraudsters and the ``early detected timestamps'' to show the number of early-detected timestamps of fraudsters.

{\noindent \bf Repeatability.} Our software together with the datasets are available at https://github.com/PanpanZheng/SAFE.

\begin{table}[]
\centering
\caption{The average performance of fraud early detection on the twitter and wiki datasets given the first 5-timestamps}
\resizebox{0.48\textwidth}{!}{
\begin{tabular}{|c|c|c|c|c|c|}
\hline
Dataset          & Algorithm & Precision     & Recall        & F1            & Accuracy      \\ \hline
\multirow{3}{*}{twitter}
                    & SVM    & $0.7370$ & $0.2733$ & $0.3875$ & $0.5916$ \\ \cline{2-6}
                    & CPH    & $0.4594$ & $\bm{0.7410}$ & $0.5440$ & $0.5453$ \\ \cline{2-6}
                    & M-LSTM    & $0.6336$ & $0.3521$ & $0.4400$ & $0.5755$ \\ \cline{2-6}
                    & SAFE     & $\bm{0.8198}$ & $0.5569$ & $\bm{0.6537}$ & $\bm{0.7180}$\\\hline \hline
\multirow{3}{*}{wiki}
                    &SVM    & $0.5484$ & $0.6413$ & $0.5911$ & $0.6754$ \\ \cline{2-6}
                    &CPH    & $0.5557$ & $0.6206$ & $0.5784$ & $0.6679$ \\ \cline{2-6}
                    &M-LSTM    & $0.5255$ & $\bm{0.9044}$ & $0.6556$ & $0.5528$ \\ \cline{2-6}
                    &SAFE     & $\bm{0.7114}$ & $0.8798$ & $\bm{0.7866}$ & $\bm{0.7640}$\\\hline
\end{tabular}
}
\label{tbl:twitter_wiki_average}
\end{table}

\begin{table*}[]
\tiny
\centering
\caption{Experimental results (mean$\pm$std.) of fraud early detection on the twitter dataset at the first 5-timestamps}

\begin{tabular}{|c|c|c|c|c|c|}
\hline
Timestamp           & Algorithm & Precision     & Recall        & F1            & Accuracy      \\ \hline
\multirow{3}{*}{@1}
                    & SVM    & $0.7500\pm0.0000$ & $0.2050\pm0.0000$ & $0.3220\pm0.0000$ & $0.5683\pm0.0000$ \\ \cline{2-6}
                    & CPH    & $0.1333\pm0.0000$ & $0.0035\pm0.0000$ & $0.0069\pm0.0000$ & $0.4901\pm0.0000$ \\ \cline{2-6}
                    & M-LSTM    & $0.6307\pm0.1072$ & $0.2350\pm0.1174$ & $0.3211\pm0.1374$ & $0.5483\pm0.0331$ \\ \cline{2-6}
                    & SAFE      & $\bm{0.8312\pm0.0313}$ & $\bm{0.3731\pm0.0987}$ & $\bm{0.5053\pm0.0870}$ & $\bm{0.6495\pm0.0309}$\\\hline \hline
\multirow{3}{*}{@2}
                    & SVM    & $0.7260\pm0.0000$ & $0.1906\pm0.0000$ & $0.3019\pm0.0000$ & $0.5593\pm0.0000 $\\ \cline{2-6}
                    & CPH   & $0.6166\pm0.0000$ & $\bm{0.7971\pm0.0000}$ & $\bm{0.6953\pm0.0000}$ & $0.6508\pm0.0000$\\ \cline{2-6}
                    & M-LSTM    & $0.6291\pm0.0734$ & $0.2952\pm0.0500$ & $0.3962\pm0.0424$ & $0.5584\pm0.0300 $\\ \cline{2-6}
                    & SAFE      & $\bm{0.8265\pm0.0297}$ & $0.5206\pm0.0564$ & $0.6360\pm0.0362$ & $\bm{0.7070\pm0.0154}$\\ \hline \hline
\multirow{3}{*}{@3}
                    & SVM    & $0.7473\pm0.0000$ & $0.2553\pm0.0000$ & $0.3806\pm0.0000$ & $0.5845\pm0.0000 $\\ \cline{2-6}
                    & CPH    & $0.5309\pm0.0000$ & $\bm{0.9389\pm0.0000}$ & $0.6783\pm0.0000$ & $0.5547\pm0.0000 $\\ \cline{2-6}
                    & M-LSTM    & $0.6239\pm0.0479$ & $0.3579\pm0.0458$ & $0.4515\pm0.0360$ & $0.5720\pm0.0223 $\\ \cline{2-6}
                    & SAFE      & $\bm{0.8193\pm0.0267}$ & $0.6016\pm0.0260$ & $\bm{0.6929\pm0.0133}$ & $\bm{0.7361\pm0.0089}$\\ \hline \hline
\multirow{3}{*}{@4}

                    & SVM    & $0.6463\pm0.0000$ & $0.1906\pm0.0000$ & $0.2944\pm0.0000$ & $0.5431\pm0.0000 $\\ \cline{2-6}
                    & CPH    & $0.5112\pm0.0000$ & $\bm{0.9820\pm0.0000}$ & $0.6724\pm0.0000$ & $0.5215\pm0.0000 $\\ \cline{2-6}
                    & M-LSTM    & $0.6256\pm0.0387$ & $0.3988\pm0.0600$ & $0.4837\pm0.0435$ & $0.5822\pm0.0200 $\\ \cline{2-6}
                    & SAFE      & $\bm{0.8136\pm0.0237}$ & $0.6330\pm0.0322$ & $\bm{0.7111\pm0.0168}$ & $\bm{0.7456\pm0.0108}$\\ \hline \hline
\multirow{3}{*}{@5}
                    & SVM    & $\bm{0.8156\pm0.0000}$ & $0.5251\pm0.0000$ & $0.6389\pm0.0000$ & $0.7032\pm0.0000 $\\ \cline{2-6}
                    & CPH    & $0.5050\pm0.0000$ & $\bm{0.9838\pm0.0000}$ & $0.6674\pm0.0000$ & $0.5098\pm0.0000 $\\ \cline{2-6}
                    & M-LSTM    & $0.6591\pm0.0547$ & $0.4739\pm0.0793$ & $0.5477\pm0.0583$ & $0.6167\pm0.0374 $\\ \cline{2-6}
                    & SAFE      & $0.8084\pm0.0424$ & $0.6564\pm0.0337$ & $\bm{0.7235\pm0.0160}$ & $\bm{0.7519\pm0.0107}$\\ \hline
\end{tabular}
\label{tbl:twitter_first_5}
\end{table*}

\subsection{Experimental Results}

\subsubsection{Fraud early detection.}

Table \ref{tbl:twitter_wiki_average} shows the average of metrics of SAFE and baselines for fraud early detection on \textit{twitter} and \textit{wiki} from @1 to @5. It is easily observed that SAFE significantly outperforms three baselines: on \textit{twitter}, accuracies and F1 scores of three baselines are all under 0.60 and 0.55, respectively, especially for CPH with accuracy 0.5453 and SVM with F1 0.3875, while SAFE obtains the acceptable accuracy 0.7180 and F1 0.6537; although three baselines improve their performance on \textit{wiki}, especially for SVM with accuracy 0.6754 and M-LSTM with F1 0.6556, however, SAFE is still far superior to them and achieves satisfiable accuracy 0.7640 and F1 0.7866. Noticeably, although CPH and M-LSTM achieve the best recall on \textit{twitter} and \textit{wiki} (0.7410 and 0.9044), however, they sacrifice their precisions with only 0.4594 and 0.5255 respectively, which indicate very high false positive rates; on the contrary, SAFE performs well on holding the balance between precision and recall such that it achieves precision 0.8198 and recall 0.5569 on \textit{twitter} and precision 0.7114 and recall 0.8798 on \textit{wiki}.

The reason why SAFE performs better than three baselines in early detection is owed to its early-detection-oriented loss function shown in Equation \ref{eq:loss_early}. Meanwhile, it also indicates that classification and typical survival models are not appropriate to early detection because their internal mechanisms do not support early detection.

Table \ref{tbl:twitter_first_5} shows the comparison results performed on \textit{twitter}.
In accordance with Table \ref{tbl:twitter_first_5}, generally speaking, the F1 and accuracy of SAFE and three baselines increase from @1 to @5. That is, whether for SAFE or three baselines, there is actually some improvement, more or less, in the performance of early detection as timestamp extends. Furthermore, we can also see SAFE performs significantly better than three baselines: at @1, accuracies of three baselines are all under 0.57, especially CPH with 0.49, which to some extent equals to random guess, while SAFE obtains an acceptable accuracy 0.6464 underlying a tracking sequence with a minimum length 12; until @5, SAFE's accuracy reaches 0.7519 while, except for SVM, Cox and M-LSTM have only 0.5098 and 0.6167, respectively.
Noticeably, it seems to be abnormal for  CPH's recall trend that it starts with 0.0035, then reaches 0.7971, and ends up with 0.9838. Although its recall is big enough, however, it has a random-guess precision around 0.5 which is not acceptable. Moreover, the reason why CPH's recall trend is so weird, we suspect, it is related to that, at least in first five timestamps, the hazards provided by time-series CPH are extremely uneven so that an appropriate survival threshold is unavailable to balance well between recall and precision expected in early detection.

\begin{figure}[htb]
    \centering
    \begin{subfigure}{0.35\textwidth}
    	\centering
	    \includegraphics[width=\linewidth, keepaspectratio]{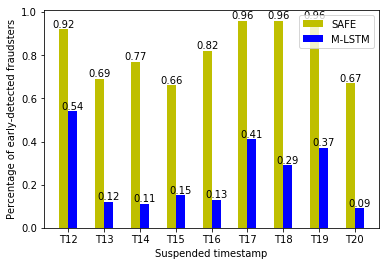}
	    \caption{Percentage of early detected fraudsters}
	    \label{fig:twitter_early_instance_number}
    \end{subfigure}
    \begin{subfigure}{0.35\textwidth}
        \centering
	    \includegraphics[width=\linewidth, keepaspectratio]{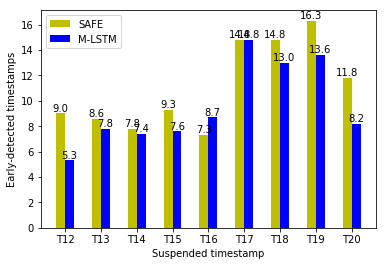}
	    \caption{Early detected timestamps of fraudsters}
	    \label{fig:twitter_early_timestamps}
    \end{subfigure}
    \caption{Comparison of SAFE and M-LSTM for fraud early detection on the twitter dataset}
\end{figure}

\subsubsection{SAFE vs M-LSTM.}
To show the advantage of survival analysis model, we further take a fine-grained comparison between SAFE and M-LSTM for  fraud early detection. M-LSTM is a classification-based model, which adopts LSTM to handle time-varying covariates. SAFE and M-LSTM have the similar neural network structure but are trained by different objective functions.  In this study, we separate all the fraudsters on \textit{twitter}  into different groups by their suspended timestamps, e.g., ``T12'' indicates the the group of fraudsters that are suspended at the 12-th timestamp. Figure \ref{fig:twitter_early_instance_number} shows the percentages of early detected fraudsters for each group by SAFE and M-LSTM. We can clearly observe that, compared with M-LSTM, SAFE has a stronger early detection capability with more early-detected fraudsters in each group. For example, at the 12-th suspended timestamp, 92\% of fraudsters are early-detected by SAFE while only 54\% of fraudsters are early-detected by M-LSTM.
 Overall, for \textit{twitter}, 82\% of fraudsters can be correctly early-detected by SAFE, while only 24\% of fraudsters can be early-detected by M-LSTM.

Figure \ref{fig:twitter_early_timestamps} shows the number of early-detected timestamps of fraudsters for each group on \textit{twitter}. We can observe that the early-detected timestamps of SAFE are still larger than those of M-LSTM in most cases. For example, for group ``T12'', SAFE can detect fraudsters with 9 timestamps ahead of the true suspended time while the early-detected timestamp of M-LSTM is 5.3. For \textit{twitter}, the average early-detected timestamp of SAFE is 11.1, while the average early-detected timestamp of M-LSTM is 9.6. Consequently, in terms of both the percentage of early-detected fraudsters and the number of early-detected timestamps, we can see SAFE obviously outperforms M-LSTM in the fraud early detection scenario.

\subsection{Model Analysis}

\begin{figure}[htb]
    \centering
    \begin{subfigure}{0.23\textwidth}
    	\includegraphics[width=\linewidth,]{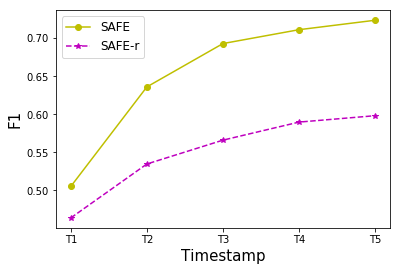}
    	\caption{F1}
    	\label{fig:f1_safd_safd_r}
    \end{subfigure}
    \begin{subfigure}{0.23\textwidth}
	    \centering
	    \includegraphics[width=\linewidth,]{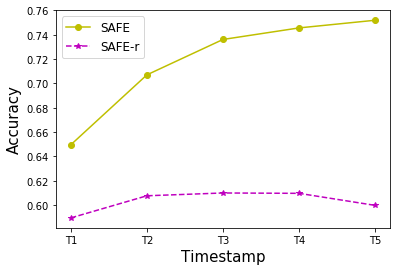}
	    \caption{Accuracy}
	    \label{fig:acc_safd_safd_r}
    \end{subfigure}
    \caption{Comparison of SAFE and SAFE-r for fraud early detection on the twitter dataset.}
    \label{fig:safd_safd_r}
\end{figure}

\subsubsection{SAFE vs. SAFE-r.}
To show the advantage of the early-detection-oriented loss function, we compare SAFE with SAFE-r that adopts regular loss function of survival analysis.
Figures \ref{fig:f1_safd_safd_r} and \ref{fig:acc_safd_safd_r} show the variation of F1 and accuracy along the timestamps on \textit{twitter}. Generally speaking, as the timestamp extends, the F1 and accuracy of SAFE and SAFE-r both increase, so their early detection performance roughly gets better. Nevertheless, we see SAFE is obviously superior to SAFE-r: from T1 to T5, the curves of SAFE for F1 and accuracy are significantly above the one of SAFE-r. Concretely, SAFE's accuracy reaches over 0.75 while SAFE-r just has 0.60 at T5. The reason behind this performance difference is associated with their loss functions. For SAFE-r, there is no internal mechanism to support it for early detection and its small performance improvement, such as accuracy from 0.57 to 0.60, is mainly due to information accumulation between steps provided by RNN; however, based on the modification of survival analysis, SAFE has its internal mechanism for early detection.

\begin{table}[]
\small
\centering
\caption{The average performance of neural survival model for fraud early detection on the twitter dataset with and without assuming prior distributions given the first 5-timestamps}
\resizebox{0.46\textwidth}{!}{
\begin{tabular}{|c|c|c|c|c|}
\hline
Algorithm & Precision     & Recall        & F1            & Accuracy      \\ \hline
					Rayleigh-RNN    & $0.5333$ & $0.0012$ & $0.0025$ & $0.5051$ \\ \hline
					Poisson-RNN     & $0.4857$ & $0.0012$ & $0.0024$ & $0.5051$ \\ \hline
          Exponential-RNN    & $0.7824$ & $0.0589$ & $0.1044$ & $0.5267$ \\ \hline
          Weibull-RNN      & $0.7381$ & $0.2865$ & $0.3850$ & $0.5920$\\\hline
                SAFE     & $\bm{0.8198}$ & $\bm{0.5569}$ & $\bm{0.6537}$ & $\bm{0.7180}$\\\hline
\end{tabular}
}
\label{tbl:twitter_first_5_ave}
\end{table}
\subsubsection{SAFE vs. Specific Distributions.}
One advantage of SAFE is that SAFE does not assume any specific distributions. We further evaluate the performance of the neural survival model with and without assuming any specific distributions. In this experiment, we train RNN to predict the parameters of a particular distribution instead of hazard rate given time-varying covariates. We adopt three common distributions for modeling the survival time, i.e., Rayleigh, Poisson, Exponential and Weibull distributions. Table \ref{tbl:twitter_first_5_ave} shows the average performance of fraud early detection on \textit{twitter} given the timestamps from 1 to 5. We can observe that SAFE, which does not assume any survival time distribution, significantly outperforms the other approaches by at least 10\% in terms of accuracy and 25\% in terms of F1. The experimental results indicate that SAFE, a model without assuming any specific distribution, is more appropriate to fraud early detection.

\section{Conclusion}
In this paper, we have developed SAFE that combines survival analysis and RNN for fraud early detection. Without assuming any fixed distribution for hazard rate, SAFE treats time-varying covariates by RNN and directly outputs hazard values at each timestamp, and then, survival probability derived from hazard values is employed to make prediction. The monotonically decreasing survival function guarantees the consistent predictions along the time.  Moreover, we revise the loss function of the regular survival model to handle training data with the late response labels. Experimental results on two real world datasets demonstrate that SAFE outperforms classification-based models, the typical survival model, and RNN-based survival models with specific distributions.
In the future, we plan to extend SAFE to predict when fraudulent activities are taken with only observed information of suspended time.

\section{Acknowledgments}
This work was supported in part by NSF 1564250 and 1841119.

\bibliographystyle{aaai}
\end{document}